%% file: paper.tex
\documentclass{jingdong}
\newcommand{\beststr}[1]{\textbf{#1}}
\newcommand{\bestall}[1]{\underline{#1}}

\definecolor{closedcolor}{RGB}{238,238,238}
\definecolor{opencolor}{RGB}{232,244,234}
\definecolor{streamcolor}{RGB}{228,239,250}
\definecolor{ourscolor}{RGB}{207,228,247}
\usepackage{float}
\usepackage{colortbl}
\usepackage{booktabs}
\usepackage{tabularx}
\usepackage{array}
\usepackage{marvosym}
\usepackage{makecell}
\usepackage{pifont}
\newcommand{\cmark}{\textcolor{red!55!black}{\ding{51}}}

\newcolumntype{Y}{>{\centering\arraybackslash}X}

\makeatletter
\newcommand{\symbolfootnotetext}[1]{%
  \begingroup
  \renewcommand{\thefootnote}{}%
  \footnotetext{\normalfont\upshape #1}%
  \addtocounter{footnote}{-1}%
  \endgroup
}
\makeatother

\title{JoyAI-Video-Edit: Real-Time Open-Ended Video Editing with Autoregressive Diffusion}

\author[\textsuperscript{*}]{Yicheng Xiao}
\author[\textsuperscript{*}]{Wenxun Dai}
\author[\textsuperscript{*}]{Xinran Qin}
\author[\textsuperscript{*}\textsuperscript{\ensuremath{\dagger}}]{Lin Song}
\author{Maoquan Zhang}
\author{Hang Xu}
\author{Yukang Chen}
\author{Yitong Li}
\author{Guohui Zhang}
\author{Yuan Zhang}
\author{Xuying Zhang}
\author{Tommy Zhang}
\author{Jianlong Yuan}
\author{Peihao Li}
\author{Shuai Lu}
\author{Siming Fu}
\author{Chuyang Zhao}
\author{Xin Han}
\author{Jie Huang}
\author{Wenbo Li}
\author{Guoqing Ma}
\author{Wei Huang}
\author{Xiaojuan Qi}
\author[\textsuperscript{\mbox{\Letter}}]{Haoyang~Huang}
\author[\textsuperscript{\mbox{\Letter}}]{Nan Duan}

\affiliation{Joy Future Academy, JD}

\input{sections/abstract}

\begin{document}

\maketitle

\symbolfootnotetext{%
  \mbox{%
    \textsuperscript{*}\,Equal contribution.
    \quad
    \textsuperscript{\ensuremath{\dagger}}\,Project leader.
    \quad
    \textsuperscript{\mbox{\Letter}}\,Corresponding author.
    \quad
    See Sec.\ref{sec:contributions} for the full author list.
  }%
}

\vspace*{-9mm}

\begin{figure}[H]
  \centering
  \includegraphics[width=0.99\textwidth]{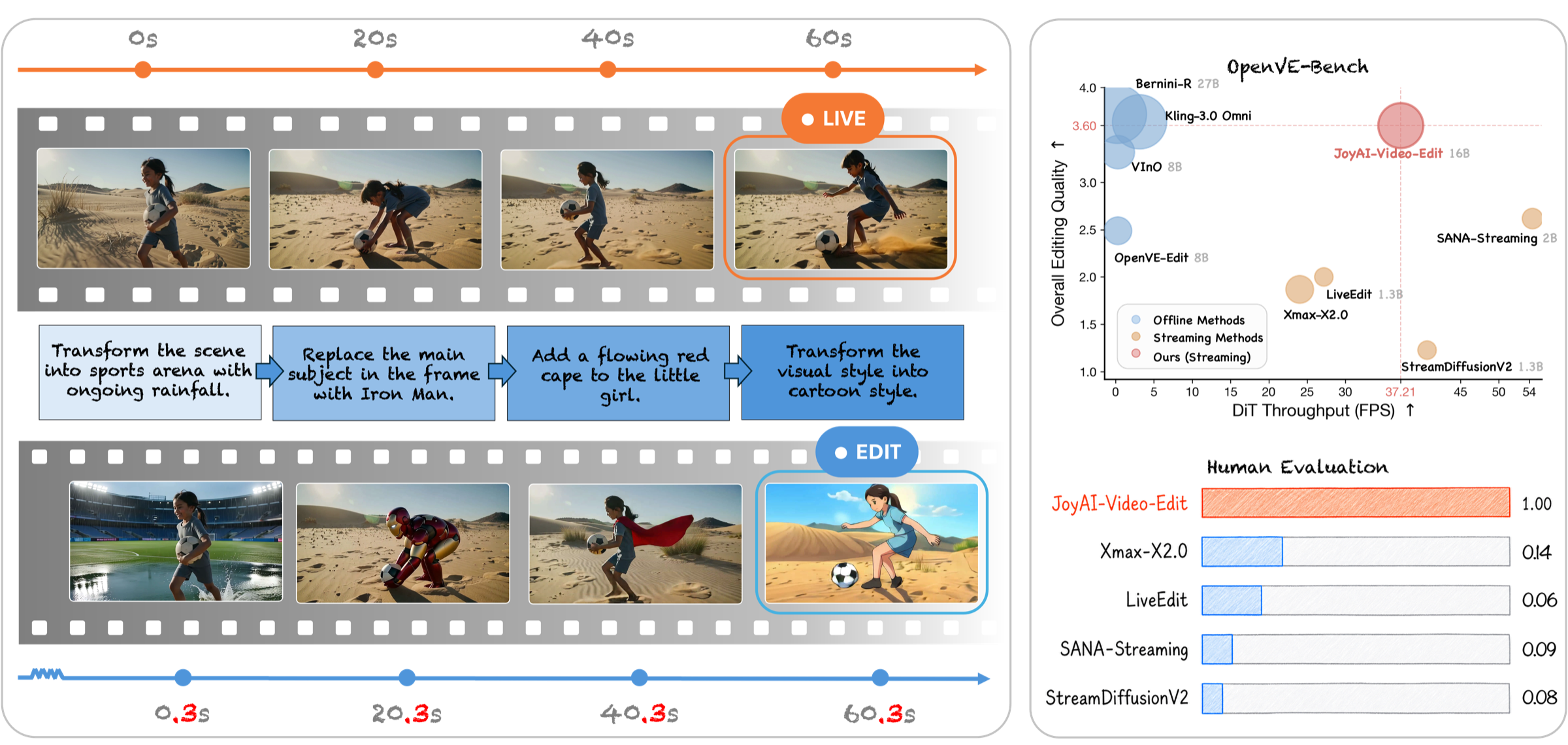}

  \caption{JoyAI-Video-Edit enables real-time, high-quality streaming
  video editing.}
  \label{fig:teaser}
\end{figure}

\begin{figure}[H]
  \centering
  \includegraphics[width=1.0\textwidth]{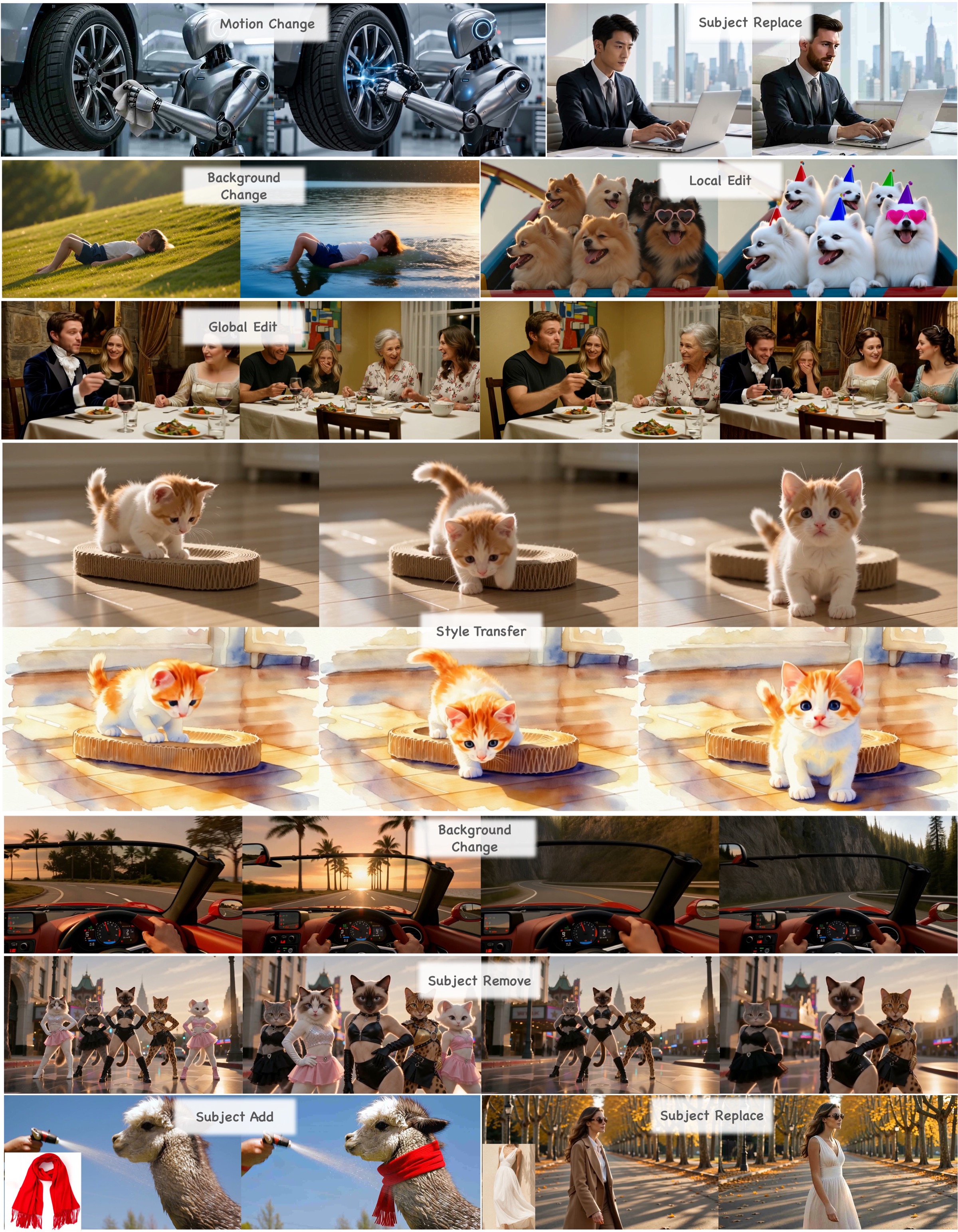}
  \vspace*{-4mm}

  \caption{JoyAI-Video-Edit supports diverse video editing tasks.}
  \label{fig:show}
\end{figure}

\input{sections/introduction}

\input{sections/related}
\input{sections/model}
\input{sections/streaming}
\input{sections/realtime}
\input{sections/data}
\input{sections/experiment}
\input{sections/acknowledge}

\clearpage

\bibliographystyle{cite}
\bibliography{main}

\end{document}

%% file: sections/abstract.tex
\abstract{
Real-time video editing requires low-latency causal generation with bounded computational resources while preserving source fidelity and long-term temporal consistency. We present JoyAI-Video-Edit, a 16B-parameter autoregressive diffusion framework for real-time, open-ended video editing without access to future frames or a predefined video duration. Our method combines chunk-wise autoregressive adaptation, Source-Anchored Distribution Matching Distillation (SA-DMD), and Long-Horizon Autoregressive Distillation to reduce train--inference mismatch, preserve source fidelity during two-step generation, and mitigate accumulated temporal drift. Extensive automatic and human evaluations show that JoyAI-Video-Edit substantially outperforms existing streaming editors and remains competitive with strong offline systems on both short and long videos. The complete system achieves end-to-end $720\mathrm{p}$ video editing at approximately 30 FPS on a single Nvidia B200 GPU.
Code is available at \url{https://github.com/jd-opensource/JoyAI-Video-Edit}.

}

%% file: sections/introduction.tex
\section{Introduction}

Instruction-guided video editing is evolving from an offline post-production tool into a continuously available visual capability.
Recent diffusion-based editors support increasingly precise and preference-aligned image manipulation~\citep{qin2025camedit,li2026hpedit}, as well as diverse video transformations including object manipulation, appearance modification, background replacement, and stylization~\citep{qi2023fatezero,jiang2025vace,ye2025unic,decart2025lucyedit,liao2025context}.
Meanwhile, efficient latent representations and selective token processing have reduced the cost of high-resolution video synthesis and localized video editing~\citep{ren2025turbo2k,wu2026yose}.
This progress opens up applications in live broadcasting, video communication, interactive entertainment, and real-time content creation, where edited frames must be produced as the source video arrives.
Unlike editing a predefined short clip, however, streaming video editing must jointly provide causal output, low response latency, bounded computation and memory, and stable editing quality over an unknown duration.

Most high-quality video editors are designed for offline, fixed-length inputs.
They process an entire clip using bidirectional or global temporal interactions and emit the result only after iterative denoising has completed~\citep{yin2025slow,wang2026liveedit,zhao2026sana}.
Although full temporal context benefits short-clip coherence, dependence on future frames prevents causal output.
Moreover, temporal tokens, attention states, and activation memory grow with the input length, while bidirectional computation limits the reuse of cached states~\citep{yang2025longlive,yuan2026helios}.
Applying an offline editor independently to consecutive clips is not a satisfactory alternative: it repeatedly processes overlapping context and can introduce visible discontinuities at clip boundaries.
A practical streaming editor must instead reuse a bounded temporal state and maintain nearly constant incremental cost as the stream grows.

Causalizing an offline editor alone does not solve this problem.
During training, an autoregressive model typically conditions on clean target history, whereas at inference time it consumes its own imperfect predictions.
This train--inference mismatch causes reconstruction errors, color deviations, and appearance changes to propagate through the generated history and accumulate into long-term drift~\citep{huang2025self,guo2025resampling}.
The challenge is particularly acute for video editing: in addition to temporal coherence, every output chunk must remain aligned with the current source chunk, preserve untargeted identity, geometry, motion, and background content, and consistently apply the requested transformation.
Generated history promotes continuity but may propagate errors, whereas the source condition preserves fidelity but may weaken editing consistency if not properly balanced.
Furthermore, the few-step generation required for real-time throughput introduces an additional gap between the iterative, classifier-free-guided training target and the single-branch model used at deployment.
Streaming video editing therefore requires causal adaptation, source-faithful few-step distillation, and explicit optimization against long-horizon error accumulation.

We present \textbf{JoyAI-Video-Edit}, a 16B-parameter autoregressive diffusion framework for real-time, open-ended video editing.
The model consists of an MLLM-based condition encoder, a causal video VAE, and a multimodal diffusion transformer, and supports both instruction-guided video-to-video editing and reference-conditioned image-and-video-to-video editing.
Starting from a strong bidirectional editor trained through a progressive image and video generation-and-editing curriculum, we convert the model into a chunk-wise causal editor.
Attention is bidirectional within each chunk and causal across chunks.
A sliding temporal window retains a fixed number of recent chunks together with the first chunk as a global sink, thereby bounding the temporal state and per-chunk computation independently of stream duration.
We first train the causal editor with clean-history teacher forcing and subsequently replace the clean history with detached, model-generated estimates through resampling forcing~\citep{guo2025resampling}.
This adaptation exposes the model to the history distribution encountered during deployment and reduces the discrepancy between training and autoregressive inference.

To enable real-time inference without sacrificing editing quality, we introduce Source-Anchored Distribution Matching Distillation (SA-DMD), which distills the iterative diffusion process into a two-step generator.
SA-DMD guides the real-score teacher independently along the text-conditioning and source-fidelity axes, using each temporally aligned source chunk to counteract drift from imperfect autoregressive history.
The source-aware guidance is applied only to the distillation target and absorbed into the generator, allowing the deployed model to preserve source fidelity with a single conditional branch.
We further introduce Long-Horizon Autoregressive Distillation, which performs segmented optimization over extended rollouts and directly supervises states affected by accumulated autoregressive errors while keeping training memory bounded.
Together, chunk-wise autoregressive adaptation, SA-DMD, and long-horizon optimization form a unified training framework that addresses causal generation, source drift, few-step acceleration, and long-term stability.
To support systematic evaluation under sustained streaming, we also construct LongV2VBench, a long-video editing benchmark covering representative global and local editing tasks.

As illustrated in Figure~\ref{fig:teaser}, JoyAI-Video-Edit supports diverse instruction-guided transformations while preserving subject identity, motion, spatial structure, and regions unrelated to the requested edit.
The edited appearance remains coherent as new source frames continuously arrive, demonstrating the model's ability to maintain both editing consistency and source fidelity over extended streams.
Extensive automatic and human evaluations show that JoyAI-Video-Edit substantially outperforms existing streaming editors and remains competitive with strong offline systems on both short- and long-video editing.
By applying SA-DMD, we reduce the diffusion denoising process to only two steps while retaining high editing quality and temporal consistency.
Together with bounded-history KV caching, FP8 quantization, and an optimized VAE pipeline, this two-step generator enables the complete system to perform end-to-end $720\mathrm{p}$ video editing at approximately 30 FPS on a single Nvidia B200 GPU.

%% file: sections/related.tex
\section{Related Work}

\subsection{Video Editing}

\textbf{Image editing foundations.}
Recent image editors have extended instruction-guided manipulation toward continuous photographic control and human-preference alignment. CamEdit enables continuous control over camera parameters, while HP-Edit introduces preference-oriented post-training for real-world editing~\citep{qin2025camedit,li2026hpedit}. These advances strengthen spatial controllability and perceptual quality, but do not address the causal temporal modeling required for streaming video editing.

\textbf{Offline video editing.}
Instruction-guided diffusion editors combine text, source videos, and optional visual references to support diverse spatial and semantic transformations~\citep{chen2026vino,lin2026kiwiedit,he2025openve,decart2025lucyedit,liao2025context,alibaba2026happyhorse,kuaishou2026kling3,bytedance2026seedance,bernini2026}.
Many high-quality systems are formulated for fixed clips and coordinate edits through bidirectional or global temporal interactions.
This formulation is effective for offline processing, whereas streaming deployment requires causal emission and reusable temporal state~\citep{yin2025slow,wang2026liveedit,zhao2026sana}.

Efficiency-oriented work reduces video diffusion cost through compressed latent representations or sparse computation. Turbo2K targets efficient high-resolution video synthesis through a highly compressed latent space, whereas YOSE selects mask-relevant tokens for efficient video object removal~\citep{ren2025turbo2k,wu2026yose}. These techniques address complementary computational bottlenecks but do not provide a general framework for open-ended instruction-guided streaming editing.

\textbf{Streaming video editing.}
Streaming methods replace full-clip processing with causal or incremental computation.
StreamDiffusionV2 develops a continuous framework for interactive video generation~\citep{feng2025streamdiffusionv2}.
SANA-Streaming subsequently combines a streaming V2V architecture, long-video training, and system optimization~\citep{zhao2026sana}, while LiveEdit transfers an offline editor to a causal model and reuses computation through an autoregressive mask cache~\citep{wang2026liveedit}.
Xmax X2.0 further targets real-time manipulation of live camera input~\citep{xmax2026x2}.
Despite enabling practical streaming, these methods often trade model capacity and computational complexity for real-time latency, limiting editing quality and versatility.
Maintaining editing consistency and instruction following over long streams also remains challenging.

\subsection{Long Video Generation}

Diffusion Forcing combines sequence prediction with diffusion by assigning independent noise levels to temporal elements~\citep{chen2024diffusion}, and subsequent systems generate videos causally from reusable or bounded histories~\citep{henschel2025streamingt2v,teng2025magi,chen2025skyreels}.
Self Forcing reduces the train--inference discrepancy by explicitly unrolling model-generated autoregressive trajectories~\citep{huang2025self}.
Because this rollout is sequential, it is formulated as post-training and relies on a few-step generator and truncated gradients to control cost, which complicates scaling to larger backbones and longer training horizons.
Resampling Forcing instead constructs model-induced histories through sequential resampling within causal diffusion training~\citep{guo2025resampling}.
Nevertheless, objectives evaluated predominantly on short trajectories may underrepresent errors that emerge only after repeated history reuse.

Few-step distillation addresses the complementary problem of inference efficiency.
Distribution Matching Distillation (DMD) compresses iterative diffusion sampling~\citep{yin2024improved} and has been adapted from bidirectional video diffusion to causal autoregressive generation~\citep{yin2025slow}.
Causal Forcing further emphasizes that distillation should account for the architectural and state-distribution mismatch between bidirectional teachers and causal students~\citep{zhu2026causal}.
For long-duration generation, LongLive-Series combines bounded causal attention with long-horizon tuning~\citep{yang2025longlive,chen2026longlive}, while Helios and Vidu S1 explore alternative routes to long-duration stability and real-time throughput~\citep{yuan2026helios,zhang2026vidus1}.
These methods cannot be directly applied to streaming video editing, which requires balancing conditioning on the source video, generated history, and current chunk to maintain long-term editing consistency. Without such a balance, errors in previously edited chunks are repeatedly propagated and accumulated throughout the autoregressive rollout.

%% file: sections/model.tex
\section{Model}
\label{sec:model}

\subsection{Model Architecture}
\label{sec:model_architecture}

\begin{figure}[t]
    \centering
    \includegraphics[width=\linewidth]{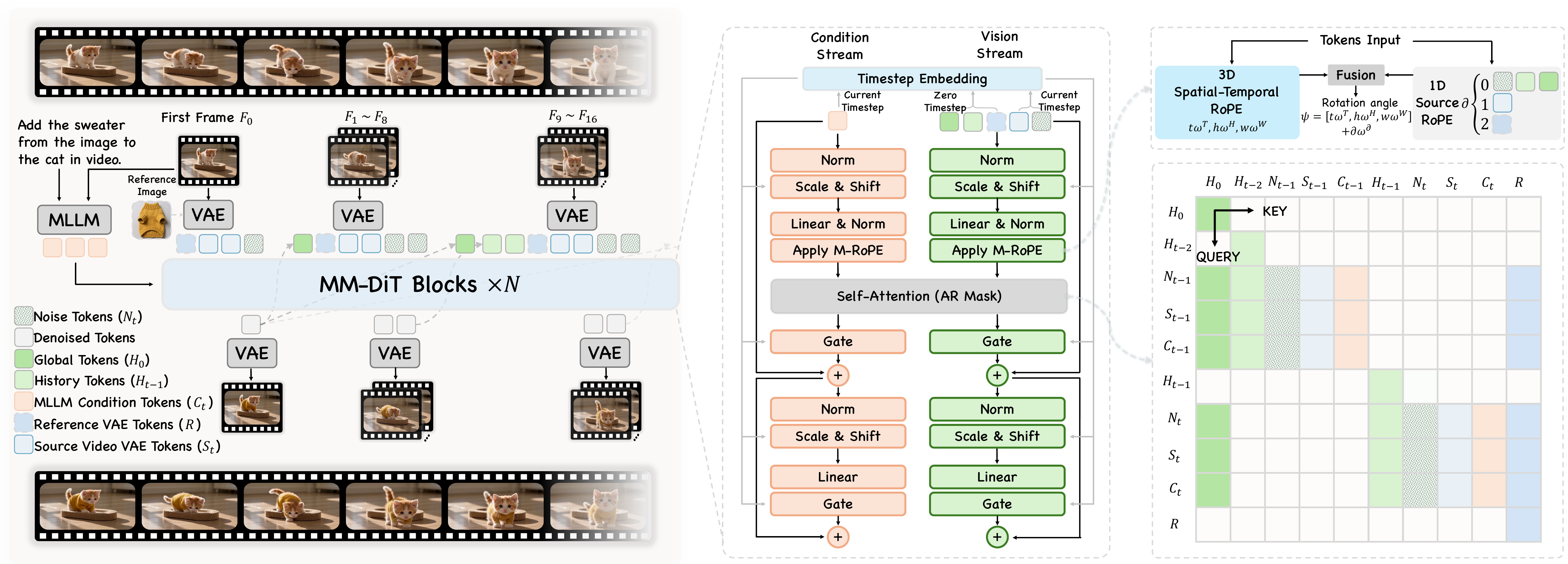}
    \caption{\textbf{Architecture of JoyAI-Video-Edit.}
    Our autoregressive diffusion architecture consists of an MLLM, a causal video VAE, and an MM-DiT diffusion backbone.
    The MLLM extracts condition tokens from textual and visual inputs, while the VAE projects videos and optional reference images into a shared latent space. The MM-DiT jointly models the condition tokens and latent visual tokens to generate edited-video latents, which are subsequently decoded by the VAE into the final output video.}
    \label{fig:model_arch}
\end{figure}

As illustrated in Figure~\ref{fig:model_arch}, JoyAI-Video-Edit is a unified autoregressive diffusion model designed for video editing. It consists of a multimodal large language model (MLLM), a causal video variational autoencoder (VAE), and a multimodal diffusion transformer (MM-DiT). Our model supports both V2V and IV2V editing under different input conditions.

\textbf{MLLM.}
Given the first frame of the source video and its corresponding editing instruction, the MLLM jointly processes the visual and textual inputs to extract condition tokens. The first frame provides the appearance and semantic context of the source video, while the instruction specifies the intended transformation. The resulting condition tokens encode both the source-aware visual information and the editing intent, providing semantic guidance to the MM-DiT throughout the denoising process.

\textbf{Causal VAE.}
The causal video VAE encodes the source video into latent sequences and maps the optional reference image into the same latent space. It uses a spatiotemporal compression ratio of $8\times24\times24$, corresponding to a temporal compression factor of $8$ and spatial compression factors of $24\times24$. Thus, each latent frame represents eight video frames.

\textbf{MLLM.}
Given the first frame of the source video and its corresponding editing instruction, the MLLM extracts condition tokens that capture the source content and the intended edit.

\begin{figure}[t]
    \centering
    \includegraphics[width=\linewidth]{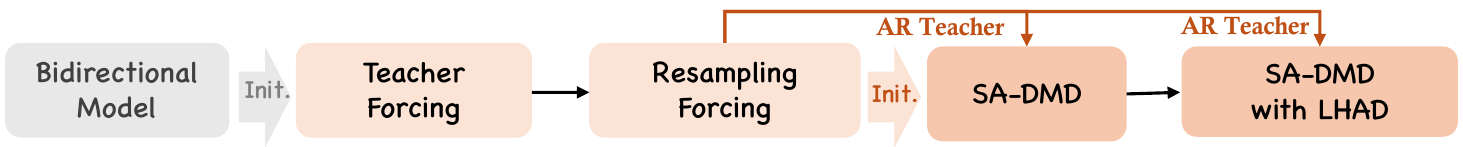}
    \caption{\textbf{The Training Pipeline of JoyAI-Video-Edit.} LHAD indicates the long-horizon autoregressive distillation.}
    \label{fig:training_pipeline}
\end{figure}

\subsection{Foundation Training}
\label{sec:t2v_foundation}

\textbf{T2V training.} We first conduct progressive T2I pretraining at resolutions of $256\times256$ and $512\times512$ to establish text--visual alignment, scene composition, and high-resolution appearance priors. The resulting T2I checkpoint is then used to initialize T2V training, during which temporal modeling is introduced while the T2I objective is retained to prevent degradation of the learned spatial generation capability.

Training begins with low-resolution videos at 12 and 24 fps and subsequently increases both the spatial resolution and frame rate. This progressive curriculum allows the model to acquire motion dynamics and long-range temporal composition without abruptly disrupting the spatial priors inherited from T2I pretraining. After large-scale pretraining, we perform supervised fine-tuning (SFT) on higher-quality data at the target resolution of $480\mathrm{p}$, followed by continual training (CT) to further consolidate visual quality, motion coherence, and prompt alignment.

\textbf{Bidirectional V2V training.}
Starting from the T2V checkpoint, we first introduce I2I supervision to learn instruction-conditioned transformation and content preservation while retaining the T2V and T2I objectives. The resulting image-editing checkpoint then initializes bidirectional video-editing training. We jointly optimize video-to-video (V2V) and image-and-video-to-video (IV2V) supervision together with the preceding tasks~\citep{jiang2025vace}. V2V extends instruction-conditioned editing over time, whereas IV2V additionally propagates appearance or identity cues from reference images.

Specifically, given a source video $x^{s}$, an editing instruction represented by the conditional tokens $c$, an optional reference image, and the edited target $x^{t}$, the causal video VAE produces the source, reference, and target latents $z^{s}$, $z^{r}$, and $z_{0}$ defined above.
The reference latent $z^{r}$ is omitted for V2V samples without a reference image.
For a sampled noise level $\sigma\in[0,1]$ and Gaussian noise $\epsilon$, we construct the noisy target and its flow target as
\begin{equation}
    z_{\sigma}=(1-\sigma)z_{0}+\sigma\epsilon,
    \qquad
    v^{\star}=\epsilon-z_{0}.
\end{equation}
The bidirectional editor $v_{\theta_{\mathrm{bi}}}$ is trained with
\begin{equation}
    \mathcal{L}_{\mathrm{V2V}}
    =
    \operatorname{E}
    \left[
        \left\|
        v_{\theta_{\mathrm{bi}}}(z_{\sigma},\sigma;c,z^{s},z^{r})
        -
        v^{\star}
        \right\|_{2}^{2}
    \right].
    \label{eq:v2v_flow_matching}
\end{equation}
Bidirectional target attention exposes the model to the complete edited clip, providing a quality-oriented editing initialization.
Bidirectional target attention establishes source preservation and temporally coherent editing before the streaming constraint is introduced. It therefore serves as the starting point for the causal autoregressive adaptation described next.

\begin{table}[t]
    \centering
    \caption{\textbf{Summary of the multi-stage foundation training curriculum.}
    Volume is measured by the number of training samples; for T2V stages, it reports the video portion of the training mixture.}
    \label{tab:foundation_training}
    \setlength{\tabcolsep}{11pt}
    \renewcommand{\arraystretch}{1.15}
    \scriptsize
    \begin{tabular}{l|c|c|c|c|c}
        \hline
        \textbf{Phase} & \textbf{Task} & \textbf{Resolution} & \textbf{fps} & \textbf{LR} & \textbf{Volume} \\
        \hline
\rowcolor{streamcolor}
\multicolumn{6}{l}{\textit{Stage 1: Text-to-Image Pretraining}} \\
        Pretrain & T2I & $256\times256$ & -- & $1\times10^{-4}$ & 4.3B \\
        Pretrain & T2I & $512\times512$ & -- & $1\times10^{-4}$ & 793M \\
        \hline
\rowcolor{streamcolor}
\multicolumn{6}{l}{\textit{Stage 2: Text-to-Video Training}} \\
        Pretrain & T2V/T2I & $256\mathrm{p}$ & 12/24 & $1\times10^{-4}$ & 370M \\
        Pretrain & T2V/T2I & $360\mathrm{p}$ & 24 & $5\times10^{-5}$ & 108M \\
        SFT & T2V/T2I & $480\mathrm{p}$ & 24 & $5\times10^{-5}$ & 30M \\
        CT & T2V/T2I & $480\mathrm{p}$ & 24 & $5\times10^{-5}$ & 6.5M \\
        \hline
\rowcolor{streamcolor}
\multicolumn{6}{l}{\textit{Stage 3: Image-Editing Training}} \\
        SFT & I2I/T2V/T2I & $720\mathrm{p}$ & - & $1\times10^{-4}$ & 3.2M \\
        \hline
\rowcolor{streamcolor}
\multicolumn{6}{l}{\textit{Stage 4: Bidirectional Video-Editing Training}} \\
        SFT & V2V/IV2V/I2I/T2V/T2I & $720\mathrm{p}$ & 24 & $5\times10^{-5}$ & 5.3M \\
        CT & V2V/IV2V/I2I/T2V/T2I & $720\mathrm{p}$ & 24 & $5\times10^{-5}$ & 1.1M \\
        \hline
    \end{tabular}
\end{table}

%% file: sections/streaming.tex
\section{JoyAI-Video-Edit}
\label{sec:joyai_video_edit}

\subsection{Chunk-wise Autoregressive Adaptation}
\label{sec:chunkwise_ar_adaptation}

We adapt the bidirectional model into a causal paradigm that generates the edited latent sequence chunk by chunk.
Each video editing pair is split along the temporal axis into aligned source and target chunks of a fixed size, with one latent frame per chunk in our implementation.
We adopt a chunk-wise attention mechanism with bidirectional attention within each chunk and causal attention across chunks.
Consequently, each target chunk can only leverage information from the current and preceding chunks, allowing the model to decode streams online without seeing future frames.
To bound the per-step computation and memory, we restrict cross-chunk attention to a sliding window.
An active chunk attends to a fixed number of recent history chunks and to the first chunk, which is retained as a global sink.
The window keeps the attention context constant regardless of video length, while the sink serves as a persistent anchor for long-horizon generation.
During training, we pack the noised active chunk together with the source tokens $S_{t}$, the condition tokens $C_{t}$, the optional reference tokens $R$, and the clean history tokens $H_{<t}$ into a single sequence, and apply an attention mask that governs the visibility of each group, as illustrated in Figure~\ref{fig:model_arch}.
The active chunk attends bidirectionally to its source and condition tokens, and causally to the in-window history and the global sink; future chunks are masked, and $R$ remains globally visible.
Following Diffusion Forcing~\citep{chen2024diffusion}, each target chunk is assigned an independent noise level during training with a masked flow-matching objective.
Teacher forcing on clean history is stable to optimize but mismatches streaming inference, where the model consumes its own imperfect and drifting predictions~\citep{huang2025self}.
Following Resampling Forcing~\citep{guo2025resampling}, we replace the clean history with an on-policy estimate, where each historical chunk is regenerated via a single-step denoising rollout and detached from gradient computation.
As a result, training is performed under a history distribution closer to that encountered during inference, alleviating the train–test distribution mismatch to some extent.

\subsection{Source-Anchored Distribution Matching Distillation}
\label{sec:aligned_ar_dmd}

To achieve real-time throughput, we distill the model into a few-step generator using Distribution Matching Distillation (DMD)~\citep{yin2024improved} within an autoregressive framework. 
The system comprises a causal generator $G_{\theta}$, a trainable fake-score model $F_{\psi}$, and a frozen real-score model $R_{\phi}$. 
Operating under a shared-backbone LoRA configuration, all three models are initialized from the weights of the previous stage.
This standard DMD objective pulls the student toward the teacher via a mode-seeking reverse-KL divergence, guided by the discrepancy between real and fake scores.
However, during extended rollouts, the generator increasingly relies on its own imperfect history, causing errors to compound into source drift and hallucinations. 
To mitigate this, we introduce {Source-Anchored} DMD (SA-DMD), which anchors the teacher to the temporally aligned source chunk. 
Specifically, we apply classifier-free guidance (CFG) to the real score along the independent axes of text condition and source fidelity:
\begin{equation}
    v_{\phi}^{g}
    =
    v_{\phi}^{\mathrm{cond}}
    +
    w_{\mathrm{txt}}\big(v_{\phi}^{\mathrm{cond}}-v_{\phi}^{-\mathrm{txt}}\big)
    +
    w_{\mathrm{src}}\big(v_{\phi}^{\mathrm{cond}}-v_{\phi}^{-\mathrm{src}}\big),
    \label{eq:source_anchored_score}
\end{equation}
where the source-free prediction $v_{\phi}^{-\mathrm{src}}$ omits the aligned source latent $S^{k}$, and $v_{\phi}^{-\mathrm{txt}}$ omits the text condition. 
This formulation renders the guided teacher a source-sharpened posterior, with $w_{\mathrm{src}}$ acting as a hyperparameter to balance history continuity against source fidelity.
Following the DMD formulation, we perturb the rollout to noise level $\sigma$ and convert the guided real and fake velocities into clean data predictions. 
The generator is then updated to minimize the normalized difference between these predictions, effectively pulling the generated distribution toward the source-anchored target. 
Concurrently, the fake score is trained via flow-matching regression on the detached generator samples.
By restricting source-anchoring strictly to the training target, the source-fidelity control is distilled directly into the generator. Consequently, the deployed model achieves high fidelity via a single conditional forward pass.

\subsection{Long-Horizon Autoregressive Distillation}
\label{sec:long_horizon_ar_distillation}

Short rollouts fail to capture the compounded errors typical of long-horizon inference, which relies heavily on self-generated history~\cite{yang2025longlive}. 
To expose the distillation to these deep states without incurring out-of-memory (OOM) errors from retaining the full computational graph, we perform segmented optimization over extended $m$-chunk rollouts. 
Specifically, we divide the sequence into shorter consecutive clips, compute the SA-DMD backward pass per clip, and clear the graph before generating the next. 
Gradients are accumulated across clips for a single optimizer step, successfully reflecting the full horizon's gradient within a bounded memory footprint. 
When target rollouts exceed the available source video length, we extend the conditioning through a dynamic {mirror looping} strategy (alternating forward and reversed sequences). 
This method helps preserve temporal continuity without materializing duplicated tensors, thereby avoiding the abrupt semantic shifts often caused by simple cyclic repetition. 
Furthermore, bounded-window causal attention ensures that KV-cache usage and per-chunk compute remain constrained during deployment, allowing the model to scale efficiently to longer sequences.

%% file: sections/realtime.tex
\section{Real-Time Deployment}
\label{sec:deployment}

At deployment, the incoming video stream is divided into consecutive eight-frame chunks $C_t$. To minimize inference overhead, we employ FP8 quantization, operator fusion, and computation-graph compilation throughout the deployment pipeline. Each chunk is encoded by the causal VAE, edited by the few-step DiT using the current condition and cached history, and immediately decoded without waiting for future frames. After generation, the clean key-value (KV) states are cached for subsequent chunks, while pseudo encoder provides the single context frame in the VAE's $(1+8)$-frame formulation. The cache retains the first chunk as a global sink and a sliding window of recent chunks, bounding both memory and per-chunk computation for open-ended streams.

\begin{figure}[t]
    \centering
    \includegraphics[width=\linewidth]{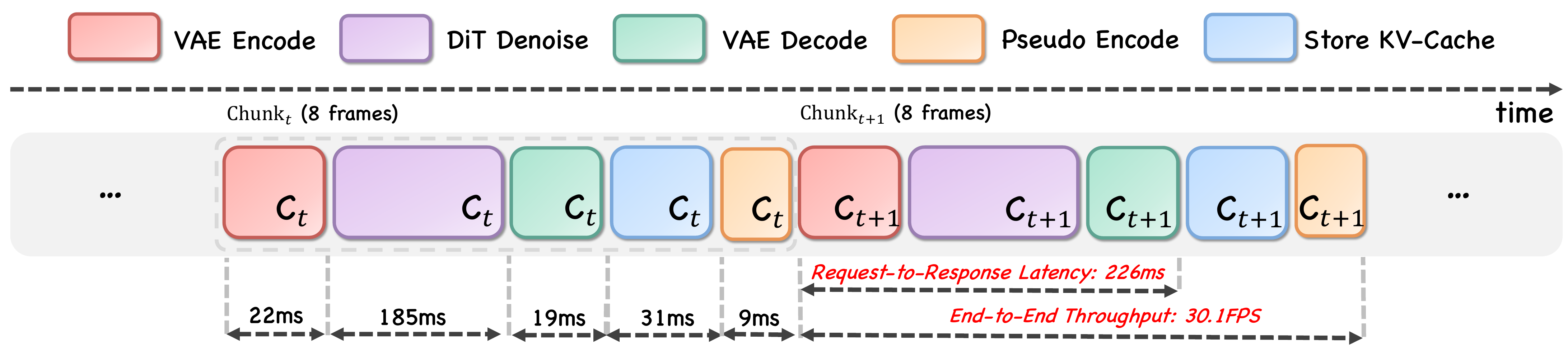}
    \caption{\textbf{Runtime analysis of JoyAI-Video-Edit on an Nvidia B200 GPU.}}
    \label{fig:runtime_pipeline}
\end{figure}

Figure~\ref{fig:runtime_pipeline} shows the stage-wise runtime on a single Nvidia B200 GPU. VAE encoding, DiT denoising, and VAE decoding take 22, 185, and 19~ms per chunk, respectively, yielding a request-to-response latency of 226~ms. Clean KV-cache construction and pseudo encoding add 31 and 9~ms, making the complete 266-ms cycle equivalent to 30.1 FPS. Reduced-precision execution, compiled and autotuned VAE paths, startup warm-up, and memory reuse further reduce runtime overhead. Together with few-step distillation and bounded KV reuse, the 16B model supports real-time $720\mathrm{p}$ editing at approximately 30 FPS on a single Nvidia B200 GPU.

%% file: sections/data.tex
\section{Data}

\begin{figure}[h]
\begin{center}
   \includegraphics[width=0.7\linewidth]{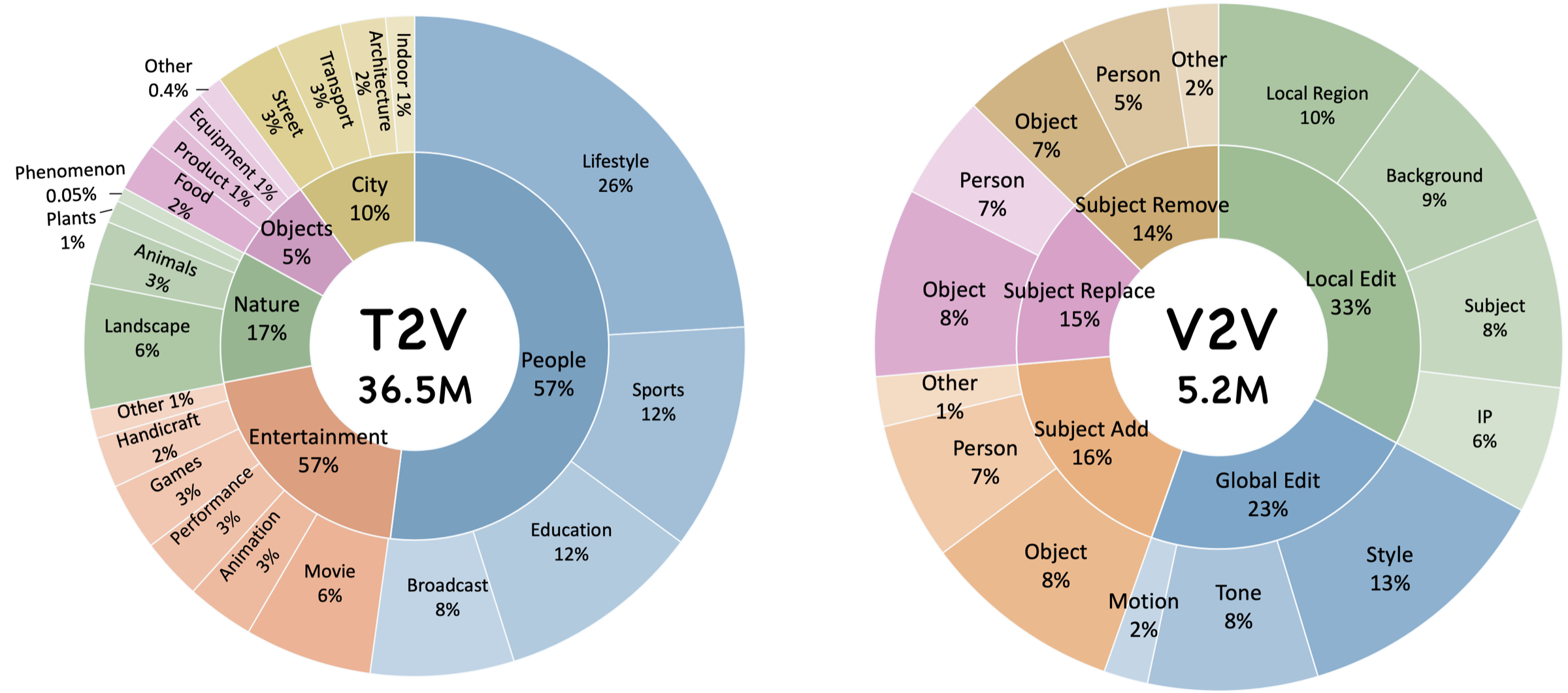}
  \caption{\textbf{Data distributions of T2V continual training (CT) and supervised fine-tuning (SFT), and V2V supervised fine-tuning (SFT).} The T2V data cover diverse semantic domains, while the V2V data include global, local, and subject-level editing tasks.}
   \label{fig:data}
\end{center}
\end{figure}

\noindent\textbf{Text-to-Video Data.}
We construct the T2V training corpus from diverse image and video sources covering people, lifestyle, entertainment, nature, objects, and urban scenes, as illustrated in Figure~\ref{fig:data}.
For images, we apply sharpness and quality assessment, black-border detection, and both perceptual-hash- and embedding-based deduplication.
For videos, we further filter samples according to visual quality, aesthetic score, motion magnitude, camera stability, and temporal validity, while removing blurred, rotated, presentation-style, and otherwise corrupted videos.

To avoid overrepresenting frequent concepts, we cluster the filtered data and downsample dominant categories while retaining long-tail concepts as much as possible.
Since aesthetic filtering may favor static videos, motion-based filtering is additionally applied to preserve samples with meaningful and stable dynamics.
For high-quality fine-tuning data, automatic filtering is followed by manual inspection to remove videos containing prominent text, broken visual structures, implausible motion, or violations of basic physical consistency.

\noindent\textbf{Video Editing Data.}
High-quality paired video-editing data are substantially more difficult to collect at scale.
We therefore transfer mature image-editing supervision to video editing using image-to-image (I2I) and reference-to-image (R2I) data from JoyAI-Image~\citep{song2026joyaiimage}.

\begin{figure}[t]
\begin{center}
   \includegraphics[width=1\linewidth]{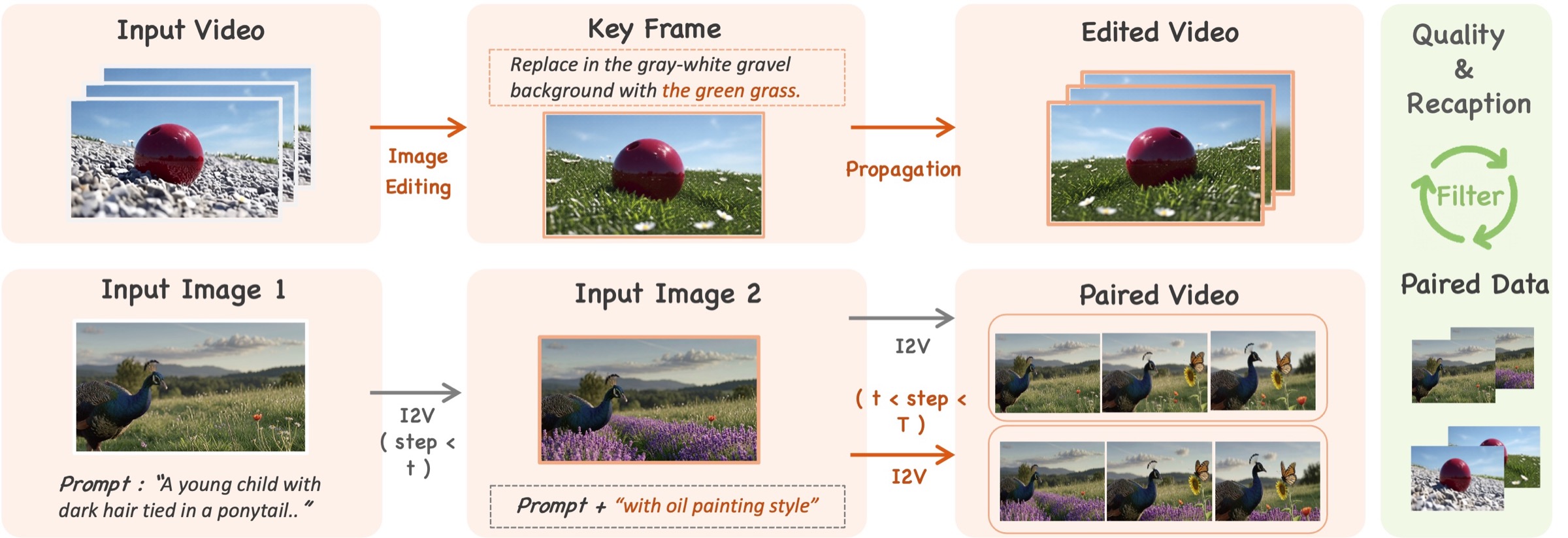}
   \caption{\textbf{Paired video editing data pipeline.}
   Paired videos are synthesized through keyframe-guided edit propagation or latent-shared I2V generation, and are subsequently filtered and recaptioned.}
   \label{fig:data_pipeline}
\end{center}
\end{figure}

As illustrated in Figure~\ref{fig:data_pipeline}, paired editing videos are constructed through two complementary pipelines.
First, a representative keyframe is selected from the source video and edited according to the instruction.
The edited keyframe and source video are then provided to an image-and-video-to-video model, which propagates the edit across the video while preserving the original motion and unedited content.
Second, paired videos can be generated from an original image and its edited counterpart using latent-shared I2V generation.
The two branches share early denoising latents to maintain consistent motion and composition, and are conditioned on different images during later denoising to introduce the desired edit.

The generated pairs are filtered according to visual quality, editing correctness, content preservation, and temporal consistency.
An MLLM subsequently compares the source and edited videos and refines their editing instructions.
As shown in Figure~\ref{fig:data}, the resulting data cover local and global edits, as well as subject addition, replacement, and removal.
Local edits mainly involve subjects, backgrounds, and specific regions, whereas global edits include style, tone, and motion transformations.


%% file: sections/experiment.tex
\section{Experiments}

\subsection{Evaluation Setup}

We compare JoyAI-Video-Edit with both streaming and offline video editors. The streaming baselines include StreamDiffusionV2~\cite{feng2025streamdiffusionv2}, SANA-Streaming~\cite{zhao2026sana}, LiveEdit~\cite{wang2026liveedit}, and the closed-source XMax-X2.0~\cite{xmax2026x2}. The offline baselines comprise the open-source VACE~\cite{jiang2025vace}, OpenVE-Edit~\cite{he2025openve}, UniVideo~\cite{wei2025univideo}, OmniWeaving~\cite{pan2026omniweaving}, Kiwi-Edit~\cite{lin2026kiwiedit}, VInO~\cite{chen2026vino}, and Bernini-R~\cite{bernini2026}, together with the commercial systems PixVerse V6~\cite{pixverse2026v6}, Runway Aleph~\cite{runwayaleph1}, Kling-3.0 Omni~\cite{kuaishou2026kling3}, and Kling-O1~\cite{kuaishou2025klingo1}.

\subsection{Automatic Evaluation}

\noindent\textbf{Short-video editing.}
We first evaluate short-video editing on the open-source OpenVE-Bench~\cite{he2025openve}. Following its five-category V2V protocol, a Gemini multimodal judge scores each edited video from 1 to 5 in terms of global style, local change, background change, local removal, and local addition. We compare against both streaming editors and strong offline systems, with results summarized in Table~\ref{tab:video-editing-comparison}.

\begin{table}[t]
\centering
\caption{Quantitative comparison of closed-source, open-source, and streaming video editing methods on OpenVE-Bench~\cite{he2025openve}. Bold: best among streaming methods; underline: best among all methods.}
\label{tab:video-editing-comparison}
\setlength{\tabcolsep}{6.0pt}
\renewcommand{\arraystretch}{1.12}
\resizebox{\textwidth}{!}{%
\begin{tabular}{lcccccccc}
\toprule
\textbf{Method} & \textbf{Params.} & \textbf{Resolution} & \textbf{Overall}
& \makecell{\textbf{Global}\\\textbf{Style}}
& \makecell{\textbf{Local}\\\textbf{Change}}
& \makecell{\textbf{Background}\\\textbf{Change}}
& \makecell{\textbf{Local}\\\textbf{Remove}}
& \makecell{\textbf{Local}\\\textbf{Add}} \\
\midrule
\rowcolor{closedcolor}\multicolumn{9}{l}{\textbf{Closed-source Methods}} \\
PixVerse V6~\cite{pixverse2026v6} & \textemdash & $720{\times}1280$ & 3.05 & 3.02 & 4.10 & 2.23 & 2.82 & 3.09 \\
Runway-Aleph~\cite{runwayaleph1} & \textemdash & $720{\times}1280$ & 3.45 & 2.62 & 4.18 & \bestall{4.16} & 2.78 & 3.49 \\
Kling-3.0 Omni~\cite{kuaishou2026kling3} & \textemdash & $1080{\times}1920$ & 3.64 & 4.03 & 4.15 & 3.20 & 3.46 & 3.36 \\
Kling-O1~\cite{kuaishou2025klingo1} & \textemdash & $1080{\times}1920$ & 3.62 & 3.38 & 4.44 & 3.23 & 3.32 & \bestall{3.74} \\
\midrule
\rowcolor{opencolor}\multicolumn{9}{l}{\textbf{Open-source Methods}} \\
VACE~\cite{jiang2025vace} & 14B & $720{\times}1280$ & 1.57 & 1.49 & 1.55 & 2.07 & 1.46 & 1.26 \\
OpenVE-Edit~\cite{he2025openve} & 5B & $704{\times}1280$ & 2.49 & 3.16 & 2.36 & 2.98 & 1.85 & 2.15 \\
Lucy-Edit~\cite{decart2025lucyedit} & 5B & $704{\times}1280$ & 2.22 & 2.27 & 1.57 & 3.20 & 1.75 & 2.30 \\
ICVE~\cite{liao2025context} & 13B & $240{\times}384$ & 2.18 & 2.22 & 1.62 & 2.57 & 2.51 & 1.97 \\
DITTO~\cite{bai2026ditto} & 14B & $480{\times}832$ & 2.13 & 4.01 & 1.68 & 2.03 & 1.53 & 1.41 \\
Kiwi-Edit~\cite{lin2026kiwiedit} & 5B & $704{\times}1280$ & \bestall{3.02} & 3.64 & 2.64 & 3.83 & 2.63 & 2.36 \\
VInO~\cite{chen2026vino} & 8B & $480{\times}848$ & 3.32 & \bestall{4.34} & 2.54 & 3.73 & 3.22 & 2.77 \\
Bernini-R~\cite{bernini2026} & 27B & $480{\times}848$ & \bestall{3.72} & 4.16 & \bestall{4.47} & 3.25 & 3.88 & 2.89 \\
\midrule
\rowcolor{streamcolor}\multicolumn{9}{l}{\textbf{Streaming Video Editing}} \\
StreamDiffusionV2~\cite{feng2025streamdiffusionv2} & 1.3B & $480{\times}832$ & 1.23 & 1.48 & 1.35 & 1.01 & 1.27 & 1.05 \\
SANA-Streaming~\cite{zhao2026sana} & 2B & $704{\times}1280$ & 2.62 & 3.48 & 2.29 & \beststr{3.20} & 2.27 & 1.88 \\
LiveEdit~\cite{wang2026liveedit} & 1.3B & $480{\times}832$ & 2.00 & 2.18 & 2.73 & 2.05 & 1.55 & 1.51 \\
Xmax-X2.0~\cite{xmax2026x2} & \textemdash & $832{\times}1440$ & 1.87 & 2.47 & 2.09 & 1.63 & 1.73 & 1.41 \\
\textbf{JoyAI-Video-Edit (Ours)} & 16B & $720{\times}1280$ & \beststr{3.60} & \beststr{3.62} & \beststr{\bestall{4.47}} & 2.90 & \beststr{\bestall{4.06}} & \beststr{2.97} \\
\bottomrule
\end{tabular}%
}
\end{table}

JoyAI-Video-Edit achieves an overall score of 3.60, outperforming SANA-Streaming~\cite{zhao2026sana}, LiveEdit~\cite{wang2026liveedit}, XMax-X2.0~\cite{xmax2026x2}, and StreamDiffusionV2~\cite{feng2025streamdiffusionv2} by 0.98, 1.60, 1.73, and 2.37 points, respectively. It ranks first among streaming methods in four of the five categories, with particularly large gains in local change and local removal. Despite causal inference, its overall performance is comparable to strong offline editors such as Kiwi-Edit~\cite{lin2026kiwiedit}, Bernini-R~\cite{bernini2026}, Kling-3.0 Omni~\cite{kuaishou2026kling3}, and Kling-O1~\cite{kuaishou2025klingo1}. It also matches the best local-change score and achieves the highest local-removal score among all compared methods, substantially narrowing the quality gap between streaming and offline video editing.

\begin{figure*}[t]
    \centering
    \includegraphics[width=\textwidth]{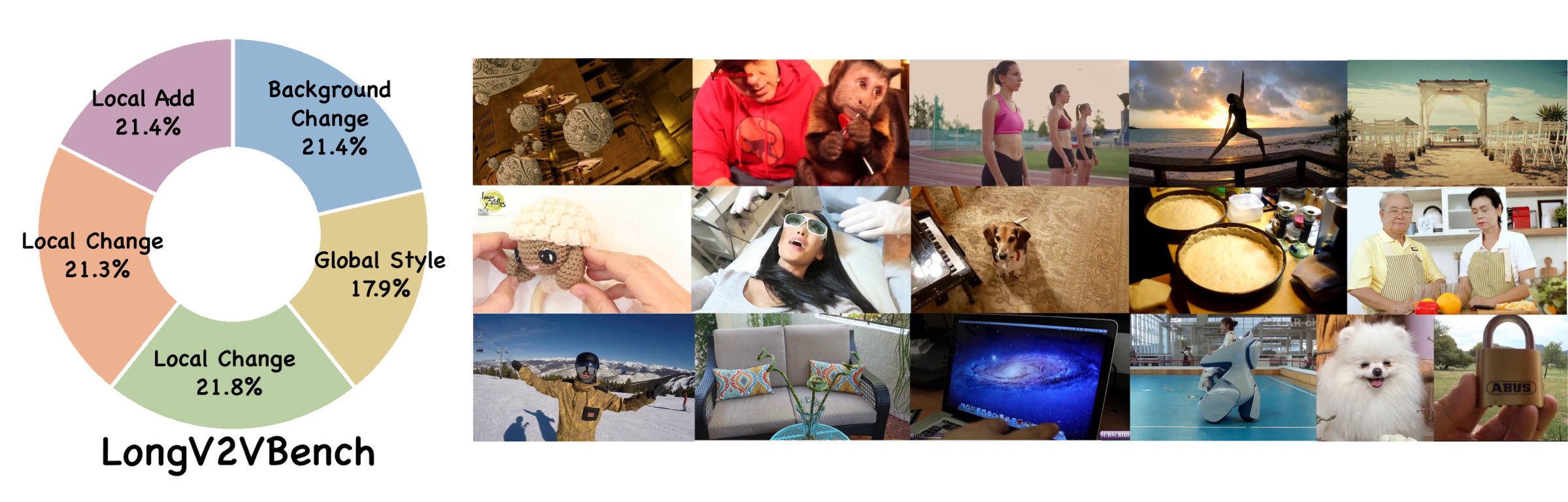}
    \caption{\textbf{Overview of LongV2VBench.}
The benchmark contains 229 long-video editing items across five categories:
background change, global style editing, local addition, local modification,
and local removal.}
    \label{fig:longbench_overview}
\end{figure*}

\noindent\textbf{Long-video editing.}
Existing video-editing benchmarks predominantly consist of clips shorter than 10 seconds, making them insufficient for assessing error accumulation, temporal degradation, and processing efficiency over sustained input streams. We therefore construct \textbf{LongV2VBench}, a one-minute video-editing benchmark comprising 229 tasks across five representative categories: background change, global style editing, local addition, local modification, and local removal. Figure~\ref{fig:longbench_overview} presents the benchmark composition and representative examples. We evaluate JoyAI-Video-Edit against existing streaming editors on LongV2VBench and report both editing quality and full-pipeline throughput.
\begin{table*}[t]
    \centering
    \caption{
        Quantitative comparison with streaming video editing methods on LongV2VBench.
        Higher editing scores indicate better editing quality.
        Full Throughput denotes the end-to-end processing speed measured in
        frames per second, where higher values indicate faster inference.
        The best result among streaming methods is shown in \textbf{bold}.}
    \label{tab:streaming-comparison}

    \setlength{\tabcolsep}{4.5pt}
    \renewcommand{\arraystretch}{1.15}

    \resizebox{\textwidth}{!}{%
    \begin{tabular}{lcccccccc}
        \toprule

        \textbf{Method}
        & \textbf{Resolution}
        & \makecell{\textbf{Throughput}\\\textbf{(FPS)} $\uparrow$}
        & \textbf{Overall} $\uparrow$
        & \makecell{\textbf{Background}\\\textbf{Change} $\uparrow$}
        & \makecell{\textbf{Global}\\\textbf{Style} $\uparrow$}
        & \makecell{\textbf{Local}\\\textbf{Add} $\uparrow$}
        & \makecell{\textbf{Local}\\\textbf{Change} $\uparrow$}
        & \makecell{\textbf{Local}\\\textbf{Remove} $\uparrow$} \\

        \midrule
StreamDiffusionV2~\cite{feng2025streamdiffusionv2}
& $480{\times}832$
& 18.07
& 1.21
& 1.08
& 1.71
& 1.18
& 1.11
& 1.03 \\

SANA-Streaming~\cite{zhao2026sana}
& $704{\times}1280$
& 14.51
& 1.64
& 1.19
& 2.02
& 1.50
& 1.72
& 1.85 \\

LiveEdit~\cite{wang2026liveedit}
& $480{\times}832$
& 15.45
& 1.23
& 1.10
& 1.34
& 1.33
& 1.34
& 1.04 \\

XMax-X2.0~\cite{xmax2026x2}
& $832{\times}1440$
& 20.90
& 1.71
& 1.36
& 2.07
& 1.64
& 2.08
& 1.40 \\

\rowcolor{streamcolor}
\textbf{JoyAI-Video-Edit (Ours)}
& $720{\times}1280$
& \beststr{30.19}
& \beststr{3.30}
& \beststr{2.49}
& \beststr{3.85}
& \beststr{3.10}
& \beststr{4.09}
& \beststr{2.99} \\

        \bottomrule
    \end{tabular}%
    }

\end{table*}

As shown in Table~\ref{tab:streaming-comparison}, JoyAI-Video-Edit achieves an overall score of 3.30 and ranks first in all five editing categories. It exceeds the strongest baseline, XMax-X2.0~\cite{xmax2026x2}, by 1.59 points overall, demonstrating that the quality advantage of our method remains consistent over one-minute videos.

Full Throughput measures the end-to-end processing speed of the complete editing pipeline. JoyAI-Video-Edit reaches 30.19 FPS at $720{\times}1280$, which is 44.4\% faster than XMax-X2.0 at a substantially lower resolution and more than twice as fast as SANA-Streaming~\cite{zhao2026sana} at a comparable resolution. This efficiency benefits from bounded-history causal inference, which reuses a fixed temporal state and maintains stable per-chunk computation as the input stream grows.

\subsection{Human Evaluation}

We conduct pairwise human evaluation between anonymized outputs from JoyAI-Video-Edit and each competing method. Evaluators select JoyAI-Video-Edit, the competing method, or a tie according to overall editing quality. All evaluation examples are single-shot videos shorter than 10 seconds.

\begin{figure}[t]
    \centering
    \includegraphics[width=\linewidth]{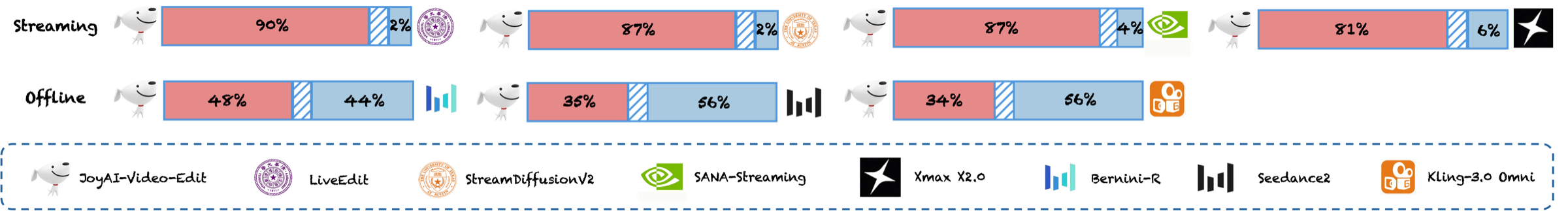}
    \caption{\textbf{Pairwise human preference between JoyAI-Video-Edit and competing methods.} Red and blue denote preferences for JoyAI-Video-Edit and the competing method, respectively; the hatched region denotes ties.}
    \label{fig:human_preference}
\end{figure}

As shown in Figure~\ref{fig:human_preference}, JoyAI-Video-Edit receives 90\%, 87\%, 81\%, and 87\% of the preference votes against LiveEdit~\cite{wang2026liveedit}, SANA-Streaming~\cite{zhao2026sana}, XMax-X2.0~\cite{xmax2026x2}, and StreamDiffusionV2~\cite{feng2025streamdiffusionv2}, respectively. Against Bernini-R~\cite{bernini2026}, JoyAI-Video-Edit receives 48\% of the votes, while Bernini-R receives 44\%. Kling-3.0 Omni~\cite{kuaishou2026kling3} and Seedance 2.0~\cite{bytedance2026seedance} each receive 56\% of the preference votes in their corresponding comparisons. Overall, JoyAI-Video-Edit is substantially preferred over existing streaming editors and achieves human-evaluated quality competitive with strong offline systems.

\subsection{Deployment Efficiency}

We further evaluate the latency and throughput of the complete deployment pipeline and its two principal neural components, the diffusion transformer and the video autoencoder. Full latency includes the end-to-end execution time of the entire editing pipeline. FPS is computed over the complete evaluated sequence and reports the mean per-frame throughput rather than instantaneous peak performance. All values are reported after deployment optimization and rounded to two decimal places.

\begin{table*}[t]
    \centering
    \caption{Latency and throughput comparison of streaming video editing methods on 81-frame inputs with batch size 1. Latency is measured over all 81 frames, and each FPS value is computed as 81 divided by the corresponding measured latency. Lower latency is better, while higher FPS is better. The best result in each column is shown in \textbf{bold}.}
    \label{tab:deployment-efficiency}
    \small
    \setlength{\tabcolsep}{2.5pt}
    \renewcommand{\arraystretch}{1.15}
    \begin{tabular*}{\textwidth}{@{\extracolsep{\fill}}lccccccc@{}}
        \toprule
        \multirow{2}{*}{\textbf{Method}}
        & \multirow{2}{*}{\textbf{Resolution}}
        & \multicolumn{2}{c}{\textbf{Full Pipeline}}
        & \multicolumn{2}{c}{\textbf{DiT}}
        & \multicolumn{2}{c}{\textbf{VAE}} \\
        \cmidrule(lr){3-4}\cmidrule(lr){5-6}\cmidrule(lr){7-8}
        & & \textbf{Latency (s)} $\downarrow$ & \textbf{FPS} $\uparrow$
        & \textbf{Latency (s)} $\downarrow$ & \textbf{FPS} $\uparrow$
        & \textbf{Latency (s)} $\downarrow$ & \textbf{FPS} $\uparrow$ \\
        \midrule
        StreamDiffusionV2~\cite{feng2025streamdiffusionv2}
        & $480{\times}832$ & 4.48 & 18.07 & 1.99 & 40.64 & 2.19 & 37.06 \\
        LiveEdit~\cite{wang2026liveedit}
        & $480{\times}832$ & 5.24 & 15.45 & 2.98 & 27.16 & 2.17 & 37.26 \\
        SANA-Streaming~\cite{zhao2026sana}
        & $704{\times}1280$ & 5.58 & 14.51 & \beststr{1.49} & \beststr{54.36} & 2.99 & 27.12 \\
        \rowcolor{streamcolor}
        \textbf{JoyAI-Video-Edit (Ours)}
        & $720{\times}1280$ & \beststr{2.68} & \beststr{30.19} & 2.18 & 37.21 & \beststr{0.405} & \beststr{200.00} \\
        \bottomrule
    \end{tabular*}
\end{table*}
As shown in Table~\ref{tab:deployment-efficiency}, JoyAI-Video-Edit achieves the lowest full-pipeline latency of 2.68 seconds and the highest end-to-end throughput of 30.19 FPS. Its full throughput is 1.67 times that of StreamDiffusionV2, 1.95 times that of LiveEdit, and 2.08 times that of SANA-Streaming. For SANA-Streaming, we benchmark its publicly available open-source deployment, which is slower than the implementation reported in the original paper~\cite{zhao2026sana}. Although SANA-Streaming has a faster standalone DiT in our evaluation, JoyAI-Video-Edit provides the best complete-system performance and reaches 200.00 FPS for VAE processing, substantially reducing the autoencoding bottleneck.

The end-to-end improvement results from the combined quantization and deployment design described in Section~\ref{sec:deployment}. Reduced-precision execution lowers the cost of both the transformer and autoencoder, while compiled VAE paths with autotuning accelerate encoding and decoding. Pipelined execution overlaps host-side processing with device computation, and startup warm-up, persisted compilation artifacts, and retained memory pools eliminate repeated compilation and allocation overhead. Together with causal KV caching that avoids recomputing previously generated content, these optimizations translate component-level acceleration into higher full-pipeline throughput.

\subsection{The Analysis of Proposed Strategies}
\input{table/ablation_algorithm}
As shown in Table~\ref{tab:ablation_algorithm}, SA-DMD drives the largest single performance boost, raising global style from 3.61 to 4.24 and local change from 3.43 to 4.00. 
This confirms that anchoring the teacher to the aligned source chunk successfully curbs the source drift that otherwise erodes appearance fidelity. 
Meanwhile, LHAD offers a complementary increase to an overall score of 3.06, particularly benefiting background change and local removal by stabilizing the late, error-accumulated states of extended rollouts. 
Ultimately, enabling both strategies yields the best overall score of 3.30 and dominates all three local-editing tasks, demonstrating that source anchoring and long-horizon optimization are highly complementary.

\begin{figure}[h]
\begin{center}
   \includegraphics[width=1\linewidth]{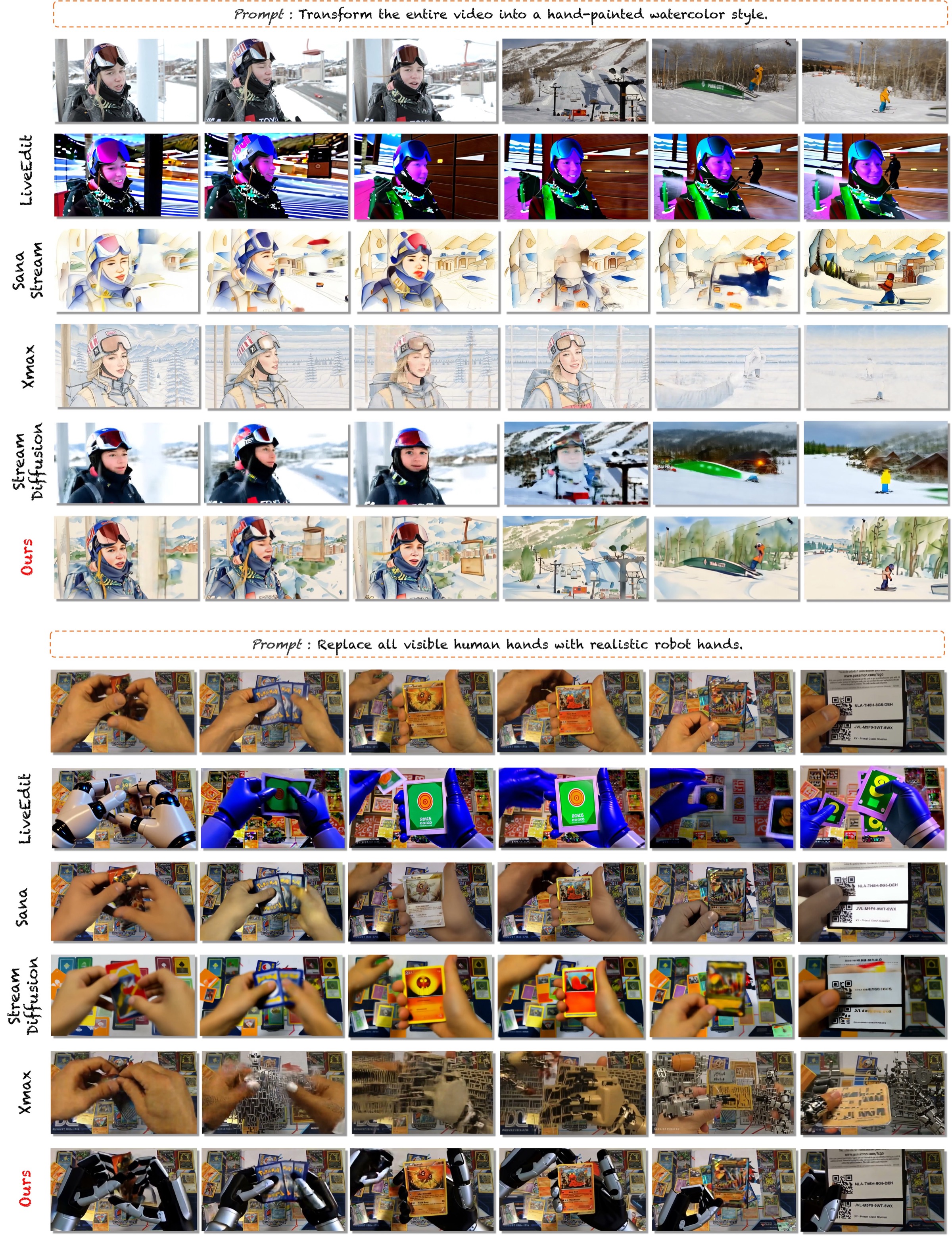}
 \caption{Qualitative comparison with streaming video editing methods.}
   \label{fig:result}
\end{center}
\end{figure}

\begin{figure}[t]
\begin{center}
   \includegraphics[width=0.96\linewidth]{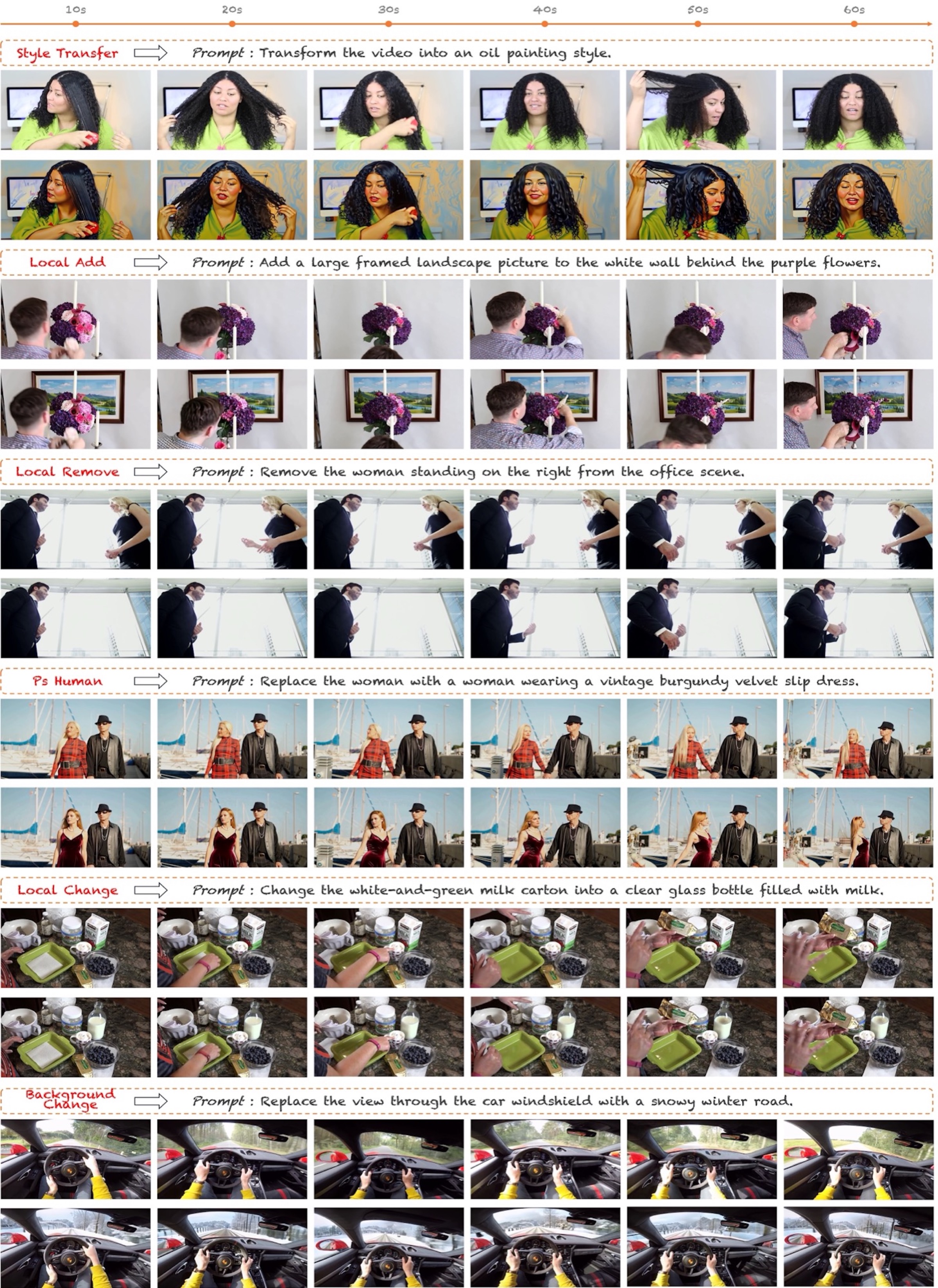}
   \caption{Additional qualitative results of JoyAI-Video-Edit across diverse editing tasks.}
   \label{fig:more_res_1}
\end{center}
\end{figure}

\begin{figure}[t]
\begin{center}
   \includegraphics[width=0.98\linewidth]{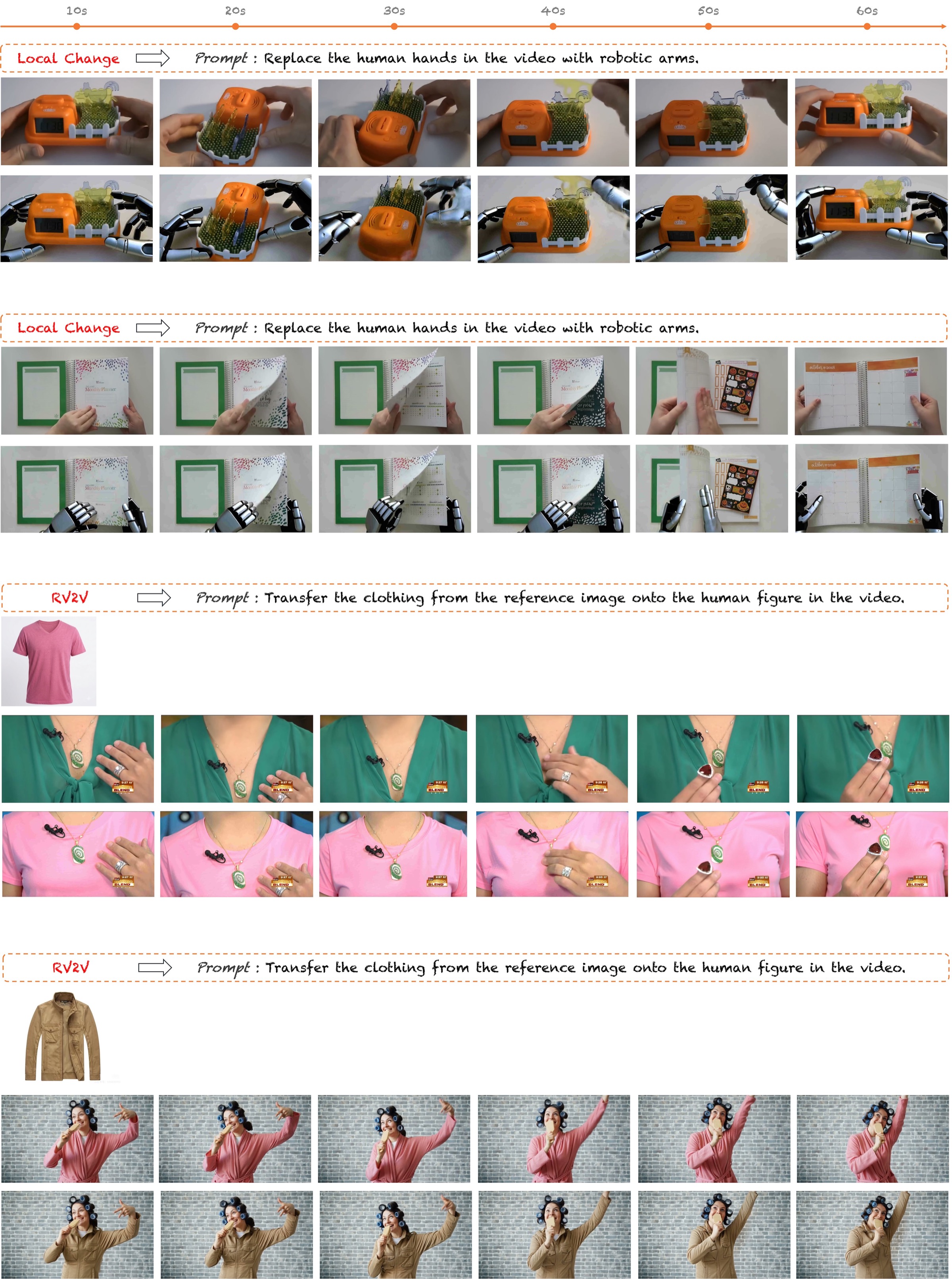}
   \caption{Additional qualitative results of JoyAI-Video-Edit across diverse editing tasks.}
   \label{fig:more_res_2}
\end{center}
\end{figure}

\subsection{Visualization Results}

JoyAI-Video-Edit transforms video editing from a fixed-clip post-production process into a continuous and interactive workflow. Figure~\ref{fig:result} illustrates its capabilities across global style transfer, scene and background replacement, localized object manipulation, appearance modification, and motion-aware editing. Across these diverse cases, the model follows the editing instruction while preserving subject identity, spatial layout, motion trajectories, and regions unrelated to the requested change. More importantly, these properties remain stable as new frames continuously arrive, allowing an edit to persist coherently over an unbounded video stream. The causal pipeline also provides immediate visual feedback, enabling users to apply or revise instructions during playback without waiting for the complete video to be processed. This capability extends video editing beyond conventional offline production to live broadcasting, real-time digital humans, interactive entertainment, game content creation, telepresence, and embodied simulation (See Figure~\ref{fig:more_res_1}). For creators, continuous preview and instruction-level control can reduce repeated rendering and shorten the iteration cycle; for platforms and enterprises, persistent stream processing makes personalized visual effects and adaptive content transformation practical at scale. JoyAI-Video-Edit therefore provides not only a faster editing tool, but also a new interaction paradigm in which video content can be modified continuously as it is captured, transmitted, or consumed.

%% file: table/ablation_algorithm.tex
\begin{table}[t]
    \centering
    \caption{\textbf{Effectiveness of the proposed distillation components on LongV2VBench.} LHAD indicates the long-horizon autoregressive distillation introduced in Section 4. Higher editing scores are better; the best result in
each column is in bold. The final row denotes our complete configuration.}
    \label{tab:ablation_algorithm}

    \setlength{\tabcolsep}{4.0pt}
    \renewcommand{\arraystretch}{1.15}

    \resizebox{0.7\linewidth}{!}{%
    \begin{tabular}{cccccccc}
        \toprule

        \textbf{SA-DMD}
        & \makecell{\textbf{LHAD}}
        & \textbf{Overall} $\uparrow$
        & \makecell{\textbf{Background}\\\textbf{Change} $\uparrow$}
        & \makecell{\textbf{Global}\\\textbf{Style} $\uparrow$}
        & \makecell{\textbf{Local}\\\textbf{Add} $\uparrow$}
        & \makecell{\textbf{Local}\\\textbf{Change} $\uparrow$}
        & \makecell{\textbf{Local}\\\textbf{Remove} $\uparrow$} \\

        \midrule

         & 
        & 2.81 & 2.45 & 3.61 & 1.97 & 3.43 & 2.58 \\

        \cmark & 
        & 3.23 & 2.49 & \textbf{4.24} & 2.74 & 4.00 & 2.67 \\

         & \cmark
        & 3.06 & \textbf{2.60} & 3.56 & 2.49 & 3.94 & 2.70 \\

        \rowcolor{streamcolor}
        \cmark & \cmark
        & \textbf{3.30} & 2.49 & 3.85 & \textbf{3.10} & \textbf{4.09} & \textbf{2.99} \\

        \bottomrule
    \end{tabular}%
}

\end{table}

%% file: sections/acknowledge.tex
\renewcommand{\thefootnote}{\fnsymbol{footnote}}

\section{Authors}
\label{sec:contributions}

\textbf{Core Contributors}\\[0.5em]
Yicheng Xiao\footnotemark[1],
Wenxun Dai\footnotemark[1],
Xinran Qin\footnotemark[1],
Lin Song\footnotemark[1]\footnotemark[2],
Maoquan Zhang,
Hang Xu,
Yukang Chen,
Yitong Li,
Guohui Zhang,
Yuan Zhang,
Xuying Zhang,
Tommy Zhang,
Jianlong Yuan,
Peihao Li,
Shuai Lu,
Siming Fu,
Chuyang Zhao,
Xin Han,
Jie Huang,
Wenbo Li,
Guoqing Ma,
Wei Huang,
Xiaojuan Qi,
Haoyang Huang\footnotemark[3],
Nan Duan\footnotemark[3]

\footnotetext[1]{Equal contribution.}
\footnotetext[2]{Project leader.}
\footnotetext[3]{Corresponding authors.}

\textbf{Contributors}\footnotemark[4]\\[0.5em]
Anson Li, Bi Cheng, Bin Li, Bo Wang, Boyang Li, Dongyan Yang, Feice Huang, Fengyuan Shi, Fan Lin, Haoran Li, Haoyu Wu, Hu Yu, Jia Shi, Jiachen Liu, Jiaqi Wang, Jiawei Li, Jianhui Liu, Jiayi Deng, Jiangmiao Pang, Junhao Zhuang, Kangliang Chen, Libing Fang, Lichen Ma, Liang Lin, Lingjie Li, Lixin Wang, Nan Jiang, Nanhua Lai, Nick, Pan Wang, Qingyi Si, Qiushi Yang, Ruofan Lv, Shaonan Wu, Tong He, Wanyan Yu, Wei Tang, Xiaoxiao Huo, Xing Pan, Xi Yang, Xuan Yang, Yan Li, Yanfei Tang, Yichen Wang, Yijun Yang, Yipeng Sun, Yuhang Li, Yujia Liang, Yue Ma, Zeyue Xue, Zuopeng Dong

\footnotetext[4]{Contributors are listed in alphabetical order.}